\documentclass{article} 
\usepackage{iclr2021_conference,times}

\usepackage[utf8]{inputenc}
\usepackage[ruled,vlined]{algorithm2e}
\usepackage{listings}
\usepackage{subfigure}
\usepackage{siunitx}
\usepackage{mathtools}
\usepackage{amsmath}
\usepackage{amssymb}
\usepackage{natbib}
\usepackage{hyperref}
\usepackage{url}
\usepackage{graphicx}

\usepackage{amsmath,amsfonts,bm}

\def\eqref#1{equation~\ref{#1}}

\def\1{\bm{1}}

\DeclareMathAlphabet{\mathsfit}{\encodingdefault}{\sfdefault}{m}{sl}
\SetMathAlphabet{\mathsfit}{bold}{\encodingdefault}{\sfdefault}{bx}{n}

\newcommand*\samethanks[1][\value{footnote}]{\footnotemark[#1]}

\title{Fair Group-Shared Representations \\ with Normalizing Flows}

\author{Mattia Cerrato\thanks{These authors contributed equally.}, Marius Köppel\samethanks[1], Alexander Segner\samethanks[1] ~\& Stefan Kramer \\
Johannes Gutenberg-Universität Mainz\\
Saarstraße 21, Mainz, Germany \\
}

\iclrfinalcopy

\begin{document}

\maketitle

\begin{abstract}
    The issue of fairness in machine learning stems from the fact that historical data often displays biases against specific groups of people which have been underprivileged in the recent past, or still are. In this context, one of the possible approaches is to employ fair representation learning algorithms which are able to remove biases from data, making groups statistically indistinguishable. In this paper, we instead develop a fair representation learning algorithm which is able to map individuals belonging to different groups in a single group. This is made possible by training a pair of Normalizing Flow models and constraining them to not remove information about the ground truth by training a ranking or classification model on top of them. The overall, ``chained'' model is invertible and has a tractable Jacobian, which allows to relate together the probability densities for different groups and ``translate'' individuals from one group to another. We show experimentally that our methodology is competitive with other fair representation learning algorithms. Furthermore, our algorithm achieves stronger invariance w.r.t. the sensitive attribute.
\end{abstract}

\section{Introduction}

The topic of fairness in machine learning has received much attention over the last years. When machine learning technologies are trained to perform prediction tasks which impact people's well-being, there is a growing concern that biased data can lead to biased decisions. For instance, the gender pay gap (see \cite{weichselbaumer2005meta}) may lead to different rates of acceptance in women's loan requests. In this situation, a financial institution that simply maximizes its utility may further perpetrate past biases based on gender. Gender, ethnicity and similar data are usually called sensitive attributes in the group fairness literature. In group fairness, one is concerned with obtaining fair results for people which belong to different groups. This property can be represented in feature space as different values of a categorical variable $s$. Various definitions of ``fair decisions'' are possible. A classifier might display \emph{disparate impact} if it takes positive decisions (e.g. getting a loan) at different rates across the groups; \emph{disparate mistreatment} instead refers to the situation where the classifier's error rate is different across the groups (see \cite{zafar2017fairness}). Another possible way to obtain fairness is learning \emph{fair representations}, i.e. representations of the original data where the groups are statistically indistinguishable from each other. Representation learning strategies that have been explored in this area include probabilistic models (see \cite{zemel2013learning}), variational models (see \cite{louizos2015vfae} and \cite{moyer2018invariant}) and reversed-gradient neural networks (see \cite{xie2017controllable}, \cite{cerrato2020constraining}, \cite{cerrato2020fairpair} and \cite{mcnamara2017provably}). The common thread between the plethora of work available in this area is information removal, i.e. employing a debiasing mechanism so that the new representation $\hat{x} = f_\theta(x)$ does not contain information about $s$. In the most general sense, perfect invariance to $s$ can be formalized as minimal mutual information between the representation and the sensitive attribute $I(\hat{x};s) = 0$ ( see \cite{moyer2018invariant} and \cite{cerrato2020constraining}).
In this work we start from the same objective of learning fair representations, but instead leverage \emph{normalizing flows}, an invertible neural network-like model that can map an arbitrary data distribution to a known one for which density estimates are trivial to compute (e.g., a Gaussian). Our fair representation algorithm learns two models, one on all the data available ($G_A$ trained on $\mathbf{x}$) and one on the data coming from a single ``pivot'' group ($G_P$ trained on $\mathbf{x_p}$); furthermore, the latent feature space $Z$ of the two models is constrained to be the same. By leveraging the invertibility of normalizing flows, our algorithm first transforms all the data into the latent feature space $Z$; then, it uses the inverse transformation $G_P^{-1}(\mathbf{z})$ to obtain \emph{shared} representations. In plain words, any individual belonging to a group $G_i$ may be ``translated'' into the feature space of pivot group $G_p$. Our approach is therefore to learn a shared feature space between groups, whereas other methods have focused on removing information about $s$ (see \cite{cerrato2020constraining}, \cite{moyer2018invariant} and \cite{louizos2015vfae}).

Our contributions may be summarized as follows:

\begin{enumerate}
    \item We present a new fair representation algorithm, FairNF, which employs normalizing flows to obtain a group-shared feature space.
    \item We test the algorithm on real world datasets in the context of fair classification and fair ranking.
    \item We discuss the actual invariance of the obtained representations.
\end{enumerate}

\section{Related Works}

For transforming data to some latent space,  normalizing flows can be used.
The learned transformation is defined as $h: Z \to X$, where $X$ is the original data and $Z$ the latent space.
One property of this mapping is that it is invertible.
As formulated by \cite{alignflow} one can use the change-of-variables formula to relate $p_X$ and $p_Z$, the marginal densities of $X$ and $Z$, respectively, by:

\begin{equation}\label{eq:change}
    p_X(x) = p_Z(h^{-1}(x)) \left | \det \frac{\partial h^{-1}(x)}{\partial x^T} \right |.
\end{equation}
Recent work has shown that such transformations can be done by deep neural networks, for example, NICE (\cite{nice}) and Autoregressive Flows (\cite{autoregressive}). 
Furthermore it was shown that evaluating likelihoods via the change-of-variables formula can be done efficiently by employing an architecture based on invertible coupling layers (\cite{dinh2017realnvp}).

In domain adaptation, the currently developed AlignFlow by \cite{alignflow}, which is a latent variable generative framework that uses normalizing flows (\cite{normflows,nice,dinh2017realnvp}), is used to transform samples from one domain to another.
The data from each domain was modeled with an invertible generative model having a single latent space over all domains.
In the context of domain adaptation, the domain class is also known during testing,  which is not the case for fairness.
Nevertheless, the general idea of training two Real NVP models is used for AlignFlow as well for the model proposed in this paper.

\section{The Fair Normalizing Flow Framework}\label{sec:model}

In this section we describe in detail our algorithm and how it can be employed to obtain fair representations. We define as $X$ the input feature space and the data vectors as $x$. In group fairness, one also has at least one variable $s$ for each data vector, representing a piece of sensitive information such as ethnicity, gender or age. This allows for the definition of a number of groups $G_1 \dots G_{\mid s \mid}$, one for each possible value of $s$. Lastly, labels $y$, representing either categorical (classification) or ordinal (ranking) information may be available in a supervised setting, which is usually the case in fair representation learning. The basic building block of our contribution is the normalizing flow model, which has been introduced in the context of generative modeling and density estimation (\cite{dinh2015nice, papamakarios2017masked, huang2018neural, dinh2017realnvp}) and since then applied to the fairness-adjacent task of domain adaptation by \cite{alignflow}. A normalizing flow model can learn a bijection $f: X \to Z$, where $Z$ is a latent feature space which may be sampled easily, such as an isotropic Gaussian distribution. Then, it is possible to estimate the density $p_X$ via $f$ and the change-of-variables formula:

\begin{align}
    p_X(x) &= p_Z(f(x)) \left | \det \frac{\partial f}{\partial x^T} \right |.
\end{align}

Our approach is to first select a pivot group $G_p$ out of the $\left | s \right |$ available ones. Then, a normalizing flow model $f_p$ is trained to learn a bijection $X_{G_p} \longleftrightarrow Z$, where we denote $X_{G_p}$ as the feature space for group $G_p$. Then, another normalizing flow model $f_{all}$ is trained independently to learn a bijection between the feature space $X$ to the same known distribution $Z$: $f_{all}:X \longleftrightarrow Z$. This makes it possible to relate the three densities $p_X$, $p_{X_{G_p}}$ and $p_Z$ to one another via the change-of-variable formula:

\begin{align}
    p_{X_{G_p}}(x_p) &= p_Z(f_p(x_p)) \left | \det \frac{\partial f_p}{\partial x_p^T} \right |, \\ 
    p_X(x) &= p_Z(f_{all}(x)) \left | \det \frac{\partial f_{all}}{\partial x^T} \right |.
\end{align}

Therefore, one may employ the two models to build a bijection chain $X \xleftrightarrow{f_{all}} Z \xleftrightarrow{f^{-1}_p} X_p$,  allowing for the transformation of vectors $x \in X$ into vectors $x_p \in X_{G_p}$. While a bijection cannot remove information about $s$, as mutual information is in general invariant to invertible transformations, the procedure is helpful in practice by obfuscating the difference between $G_p$ and $G_{all}$, as we show in Section~\ref{sec:exp_results}. 

As commonly done in the normalization flow literature, we train our models with a log likelihood loss (see, e.g.,  \cite{dinh2015nice} and \cite{papamakarios2017masked}). We rely on the ``coupling layers'' architecture, which guarantees invertibility, as introduced by \cite{dinh2017realnvp}.

To ensure that the new representations can still be used to predict the target labels $y$, a classification or ranking model can be trained on top of them.
We included a loss evaluated on the target label which is propagated through both normalizing flow models.
This leads to a two-fold loss function,  which can be regulated by the $\gamma$ hyperparameter to address the relevance-fairness trade-off:

\begin{align}
    L &= \gamma (L_{f_p} + L_{f_{all}} ) + L_y.
\end{align}
Here, $L_y$ can be any loss function which can be evaluated on $y$.
The gradients of $L_{f_p}$ and $L_{f_{all}}$ are only applied to $f_p$ and $f_{all}$, respectively, while the ones of $L_y$ are applied to $f_p$, $f_{all}$ and the model predicting the target labels $y$.

\section{Experiments}\label{sec:exp_results}

\subsection{Experimental Setup}
To overcome statistical fluctuations during the experiments, we split the datasets into 3 internal and 3 external folds.
On the 3 internal folds, a Bayesian grid search, optimizing the fairness measure, is used to find the best hyperparameter setting.
The best setting is then evaluated on the 3 external folds.

We compare our model with a state-of-the-art algorithm called Fair Adversarial DirectRanker (AdvDR in the rest of this paper),  which showed good results on commonly used fairness datasets (see \cite{cerrato2020fairpair}), a Debiasing Classifier (AdvCls) used in \cite{cerrato2020constraining} and a fair listwise ranker (DELTR, \cite{deltr}).

Since it is from a theoretical point (to the best of our knowledge) not clear how exactly a Real NVP model is treating discrete features, we designed two different experiments whether the model is able to be trained on discrete features.
For the first experiment done on the Compas dataset (see \cite{dieterich2016compas}), we included discrete and continuous features.
For the second experiment done on the Adult (see \cite{adult}) and Banks dataset (see \cite{banks}) we only used continues features. 

For our implementation of the models and the experimental setup, see~\textit{\url{https://zenodo.org/record/4566895}}.

\subsection{Experimental Results}

\begin{table}[!ht]
\centering
\begin{tabular}{lllll}
\hline
  Model & Dataset &              1-rND &              1-GPA &               NDCG@500 \\
\hline
    FairNF &  Compas &  0.838 $\pm$ 0.059 &   0.934 $\pm$ 0.030 &  0.474 $\pm$ 0.115 \\
    FairNF &   Adult &  0.922 $\pm$ 0.009 &  0.929 $\pm$ 0.025 &   0.460 $\pm$ 0.054 \\
    FairNF &   Banks &  0.859 $\pm$ 0.017 &  0.892 $\pm$ 0.016 &  0.519 $\pm$ 0.011 \\
    \hline
  AdvDR &  Compas &  0.864 $\pm$ 0.061 &  0.911 $\pm$ 0.047 &  0.424 $\pm$ 0.033 \\
  AdvDR &   Adult &  0.884 $\pm$ 0.036 &  0.975 $\pm$ 0.016 &  0.647 $\pm$ 0.087 \\
  AdvDR &   Banks &  0.778 $\pm$ 0.106 &  0.735 $\pm$ 0.036 &  0.521 $\pm$ 0.086 \\
  \hline
 AdvCls &  Compas &  0.823 $\pm$ 0.057 &  0.917 $\pm$ 0.032 &  0.542 $\pm$ 0.055 \\
 AdvCls &   Adult &  0.929 $\pm$ 0.007 &  0.901 $\pm$ 0.021 &  0.629 $\pm$ 0.008 \\
 AdvCls &   Banks &  0.918 $\pm$ 0.051 &  0.972 $\pm$ 0.026 &  0.198 $\pm$ 0.176 \\
 \hline
  DELTR &  Compas &  0.825 $\pm$ 0.072 &  0.926 $\pm$ 0.007 &  0.438 $\pm$ 0.202 \\
  DELTR &   Adult &  0.744 $\pm$ 0.087 &  0.742 $\pm$ 0.048 &  0.142 $\pm$ 0.119 \\
  DELTR &   Banks &  0.823 $\pm$ 0.057 &  0.917 $\pm$ 0.032 &  0.542 $\pm$ 0.055 \\
\hline
\end{tabular}
\caption{Results for different fairness datasets. The used fairness metrices are 1-rND and 1-GPA used by \cite{cerrato2020fairpair} and the commonly used relevance metric NDCG@500. The used datasets are commonly used in fairness. For the Compas dataset (see \cite{dieterich2016compas}) we used being black or white as sensitive attribute while we took the risk score as the ground truth. For the Adult dataset (see \cite{adult}) we used for the ground truth whether an individual’s annual salary is over 50K\$ per year or not. The sensitive attribute was being a man or a woman. For the Banks dataset (see \cite{banks}) the classification goal is whether a client will subscribe a term deposit. This dataset is biased against people under 25 or over 65 years.}
\label{tab:results}
\end{table}

In Table \ref{tab:results} the results for different fairness datasets are shown.
FairNF is the proposed algorithm explained in Section \ref{sec:model}.
In terms of the two fairness measures,  this algorithm is outperforming the other approaches in at least one measure over all datasets.
The only algorithm that is better than FairNF in terms of fairness is AdvCls on the Banks dataset.
However, on this dataset the algorithm is performing poorly on NDCG@500, while FairNF has a stable performance.
In terms of relevance, FairNF and AdvDR have similar performance on the Compas and Banks dataset, while  AdvDR is better on the Adult one.
The other two algorithms may have weaker results on relevance (AdvCls on Banks; DELTR on Adult).
Looking at the question whether the model is able to be trained on continuous features only, we cannot see any significant performance difference comparing the experiments done on Compas compared with the ones done on Adult and Banks.

\section{Conclusion and Future Works}

We introduced a new algorithm, called FairNF, which is able to transform unfair data into fair data by learning group-shared representations. This is made possible by training a pair of Real NVP models, with one focusing on a pivot group and the other learning from the whole dataset.
The models are then used to build a bijection chain, allowing to obfuscate the difference between the sensitive groups.
The two Real NVP models were constrained to still provide information about the ground truth by training a ranking or classification model on top of them and propagating the gradients throw the whole bijection chain.
The proposed algorithm performed well compared to existing methods on the tested Compas, Adult and Banks dataset.
We are currently investigating an extension of our framework in the context of learning a \emph{disentangled} latent space $Z$. A disentangled latent space would dedicate a number of dimensions to represent all information about the sensitive attribute. The remaining dimensions may then be employed to project back into the original feature space. We expect this to provide further benefits in terms of representation invariance and fairness.
Since the empirical results including both discrete and continuous features into the training data did not show any performance decrease, we would further investigate this by including a deeper theoretical analysis on this property.

\newpage

\section*{Supplementary Material}

\subsection*{The NDCG Metric}\label{sec:ndcg}
The normalized discounted cumulative gain of top-$k$ documents retrieved (NDCG@$k$) is a common used measure for performance in the field of learning to rank.
Based on the cumulative gain of top-$k$ documents (DCG@$k$) the NDCG@$k$ can be computed by dividing the DCG@$k$ by the ideal (maximum) discounted cumulative gain of top-$k$ documents retrieved (IDCG@$k$):
\begin{equation}
\text{NDCG@}k = \frac{\text{DCG@}k}{\text{IDCG@}k} = \frac{\sum_{i=1}^{k} \frac{2^{r(d_i)} - 1}{log_2(i + 1)}}{\text{IDCG@}k}
\,,\notag
\end{equation}
where $d_1, d_2, ..., d_n$ is the list of documents sorted by the model with respect to a single query and $r(d_i)$ is the relevance label of document $d_i$.

\subsection*{rND}\label{sec:rnd}

To the end of measuring fairness in our models we employ the rND metric which was introduced by~\cite{rnd}. 
This metric is used to measure group fairness and is defined as follows: 

\begin{equation}
\text{rND} = \frac{1}{Z} \sum_{i \in \{10, 20, ...\}}^N \frac{1}{log_{2}i} \mid \frac{ \mid S^{+}_{1...i} \mid}{i} - \frac{\mid S^+ \mid}{N} \mid.
\end{equation}

The goal of this metric is to measure the difference between the ratio of the protected group in the top-i documents and in the overall population.
The maximum value of this metric is given by $Z$ which is also used as normalization factor.
This value is computed by evaluating the metric with a dummy list where the protected group is placed at the end of the list. This biased ordering represents the situation of ``maximal discrimination''. 

This metric also penalizes if protected individuals at the top of the list are over-represented compared to their overall representation in the population.

\subsection*{Group-dependent Pairwise Accuracy}\label{sec:gpa}

Let $G_1, ..., G_K$ be a set of $K$ protected groups such that every document inside the dataset $D$ belongs to one of these groups. The \emph{group-dependent pairwise accuracy} introduced in \cite{fair_pair_metric} $A_{G_i > G_j}$ is then defined as the accuracy of a ranker on documents which are labeled more relevant belonging to group $G_i$ and documents labeled less relevant belonging to group $G_j$. Since a fair ranker should not discriminate against protected groups, the difference $|A_{G_i > G_j} - A_{G_j > G_i}|$ should be close to zero. In the following, we call the Group-dependent Pairwise Accuracy {\em GPA}.

\subsection*{Fair Representations}\label{sec:fair_rep}

\begin{table}[!ht]
\centering
\begin{tabular}{lcc} 
 \hline
 Method & $x$ & $\hat{x}$ \\
 \hline
 LR (ADRG)& \SI{0.095 \pm 0.014}{} & \SI{0.046 \pm 0.024}{} \\
 RF (ADRG)& \SI{0.084 \pm 0.012}{} & \SI{0.052 \pm 0.010}{} \\
 MLP (ADRG)& \SI{0.091 \pm 0.016}{} & \SI{0.067 \pm 0.013}{} \\
 
 LR (F1)& \SI{0.676 \pm 0.013}{} & \SI{0.649 \pm 0.048}{} \\
 RF (F1)& \SI{0.667 \pm 0.009}{} & \SI{0.657 \pm 0.009}{} \\
 MLP (F1)& \SI{0.681 \pm 0.009}{} & \SI{0.659 \pm 0.016}{} \\
 
 LR (AUC)& \SI{0.595 \pm 0.014}{} & \SI{0.546 \pm 0.024}{} \\
 RF (AUC)& \SI{0.584 \pm 0.012}{} & \SI{0.552 \pm 0.010}{} \\
 MLP (AUC)& \SI{0.591 \pm 0.016}{} & \SI{0.567 \pm 0.013}{} \\
 \hline
 \end{tabular}
 \caption{ADRG, F1 and AUC scores of the representation results for the original data $x$ and the transformed data $\hat{x}$ using our framework on the COMPAS dataset. ADRG stands for ``Absolute Difference from Random Guess'', i.e. the difference in accuracy from a classifier which always predicts the majority class (see \cite{cerrato2020fairpair}).}
\label{tab:rep}
\end{table}

For further investigating the fairness of the new representations we trained external classifiers on the transformed data to predict the sensitive attribute.
In Table~\ref{tab:rep} the results on the original COMPAS data ($x$) and the transformed data $\hat{x}$ are shown. 
The used classifiers are a Linear Regression (LR), Random Forest (RF) and a simple Multilayer Perceptron (MLP).
They are also trained five times over the five different train/test folds.
For the evaluation the F-score was taken which also takes unbalanced class representations into account.
A lower score means a less precise prediction.
Overall all algorithms are worse in predicting $s$ while trained with $\hat{x}$ compared to the results for the original dataset. Therefore, we observe gains in representation invariance by employing our methodology.

\subsection*{Datasets}
For evaluating on real world data the COMPAS dataset (see \cite{machine_bias}), was used.
This dataset was released as part of an investigative journalism effort in tackling automated discrimination.
Furthermore the Adult dataset is used where the ground truth represents whether an individual's annual salary is over 50K\$ per year or not (\cite{adult}). 
It is commonly used in fair classification, since it is biased against gender(see \cite{fair_autoencoder}, \cite{zemel2013learning} and \cite{cerrato2020constraining}).
The third dataset used in our experiments is the Bank Marketing Data Set (see \cite{banks}) where the classification goal is whether a client will subscribe a term deposit.
The dataset is biased against people under 25 or over 65 years.

\subsection*{Toy Data}
For evaluating the FairNF algorithm we generate two sets of toy data by randomly sampling two features from different normal distributions.
The mean and the standard deviation of each distribution is set to be different.
The toy data is then processed using our framework.
The goal for this experiment is show how the two distributions can be transformed to one.

\begin{figure*}[!ht]
\centering
\subfigure[Original toy data]{\label{fig:toy_1}\includegraphics[width=0.49\textwidth]{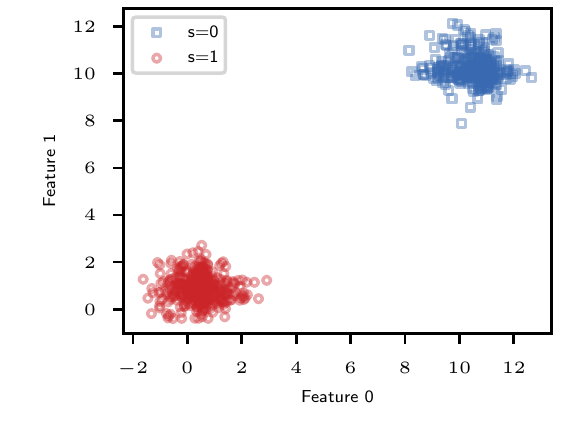}}
\subfigure[Toy data after applying the framework]{\label{fig:toy_2}\includegraphics[width=0.49\textwidth]{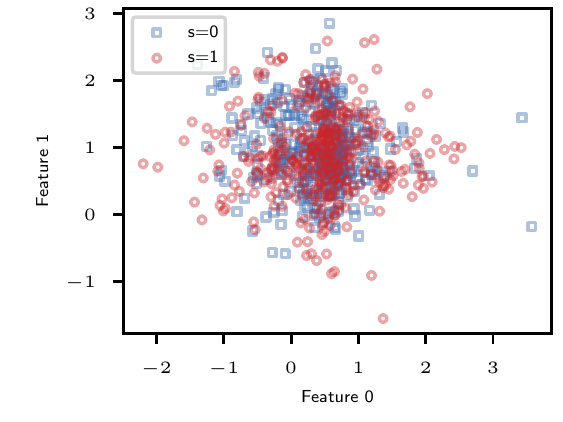}}
\caption{In~\ref{fig:toy_1} the generated toy data is shown where the different groups are displayed with blue and red. In~\ref{fig:toy_2} the transformed toy data is shown using the FairNF algorithm.}
\end{figure*}

In~\ref{fig:toy_1} the original toy data is shown while~\ref{fig:toy_2} shows the data after the transformation with the FairNF algorithm.
In~\ref{fig:toy_1} the two distributions are not overlaying so they should be easily distinguished by a classification algorithm.
After transforming them they heavily overlap making it harder to distinguish between the two.

\end{document}